\documentclass[11pt]{article}
\usepackage{array}
\usepackage{colortbl}
\usepackage[preprint]{acl}
\usepackage{times}
\usepackage{latexsym}
\usepackage[T1]{fontenc}

\usepackage[utf8]{inputenc}

\usepackage{microtype}

\usepackage{inconsolata}

\usepackage{graphicx}
\usepackage{arydshln}
\usepackage{pifont}  
\usepackage[table]{xcolor}  

\newcommand{\bibo}{{P\textsc{rism}}}
\title{\bibo: A Plug-in Reproducible Infrastructure\\ for Scalable Multimodal Continual Instruction Tuning}

\author{Jun-Tao Tang\textsuperscript{2,*},
        Yu-Cheng Shi\textsuperscript{2,*},
        Zhen-Hao Xie\textsuperscript{1,2},
        Da-Wei Zhou\textsuperscript{1,2,\textdagger} \\
        \textsuperscript{1} School of Artificial Intelligence, Nanjing University, China \\
        \textsuperscript{2} National Key Laboratory for Novel Software Technology, Nanjing University, China \\
        \textsuperscript{*} Equal contribution \\
        \textsuperscript{\textdagger} Correspondence: \href{zhoudw@lamda.nju.edu.cn}{zhoudw@lamda.nju.edu.cn}
}

\begin{document}
\maketitle
\begin{abstract}
Multimodal Large Language Models (MLLMs) achieve versatility by reformulating diverse tasks into a unified instruction-following framework via instruction tuning. However, real-world deployment requires continuous adaptation to emerging tasks, motivating Multimodal Continual Instruction Tuning (MCIT). Despite its growing importance, current MCIT research is hindered by severe engineering bottlenecks. 
Existing methods are typically implemented by directly modifying the base MLLM codebase, which imposes substantial implementation overhead and yields method-specific architectures that severely limit code reuse and fair comparison.
To address this, we introduce \bibo, a plug-in reproducible codebase specifically designed for scalable MCIT research. It separates algorithmic development from the backbone implementation via a lightweight plugin registration mechanism, enabling new strategies to be integrated as independent plugins without modifying the underlying MLLM codebase, thereby eliminating structural fragmentation and accelerating method development. \bibo\ natively supports widely used large-scale training pipeline, thereby enabling reproducible and scalable MCIT experimentation. Code is available at https://github.com/LAMDA-CL/Prism.
\end{abstract}

\section{Introduction}
Recently, multimodal Large Language Models (MLLMs)~\cite{Qwen-VL,zhu2023minigpt} have demonstrated remarkable potential across diverse domains, largely driven by their ability to interpret and execute tasks through natural language instructions. 
Through instruction tuning~\cite{zhang2023instruction,tong2025metamorph}, MLLMs reformulate both unimodal vision tasks (\textit{e.g.}, image classification and visual grounding~\cite{deng2021transvg}) and vision-language tasks (\textit{e.g.}, visual question answering~\cite{goyal2017making}) into a unified instruction-following framework~\cite{lee2024visual}, thereby achieving unprecedented versatility.
However, real-world deployment operates in dynamic environments where data arrives as a continuous stream~\cite{krempl2014open}. To maintain long-term utility, MLLMs must continuously absorb new knowledge and adapt to emerging instruction formats via continual instruction tuning. Conventional fine-tuning methods, when applied sequentially to such evolving data streams, tend to overwrite previously learned representations, resulting in catastrophic forgetting of prior capabilities~\cite{mccloskey1989catastrophic,zhou2024class}. To address this fundamental challenge, Multimodal Continual Instruction Tuning (MCIT)~\cite{chen2024coin,xie2026same} has emerged as a critical research paradigm, focusing on equipping MLLMs with the capacity to learn incrementally while rigorously preserving established knowledge.

Current MCIT research faces significant engineering challenges. Most existing methods are implemented by directly modifying the base MLLM training codebase. Given the architectural complexity of modern MLLMs, such modifications lead to highly divergent code structures and training logic across approaches. In existing toolkits~\cite{chen2024coin,guo2025mcitlib}, each method maintains a full copy of the MLLM codebase, tightly coupling algorithmic logic with core training infrastructure. Consequently, these frameworks lack a highly integrated and decoupled architecture. This structural fragmentation obscures core implementation details, making code reuse and subsequent development significantly more challenging. Furthermore, many traditional continual learning techniques do not support essential large-scale training infrastructure, such as gradient checkpointing~\cite{chen2016training} and DeepSpeed~\cite{rasley2020deepspeed}. This incompatibility severely restricts their scalability to MLLMs and hinders fair comparisons with continual learning baselines.

\begin{table}[t]
\centering
\small
\renewcommand{\arraystretch}{1.0}
\setlength{\tabcolsep}{4pt}
\caption{Comparison of \bibo\ to existing representative MCIT toolkits.}
\label{tab:comp}
\begin{tabular}{lccc}
\hline
\rowcolor{gray!15}
\textbf{Feature} & \textbf{CoIN} & \textbf{MCITlib} & \textbf{\bibo}\\
\hline
Implemented Algorithms      & 4 & 8 & 9 \\
Supported Benchmarks        & 1 & 3 & 3 \\
Unified Backbone Design            & \ding{55} & \ding{55} & \ding{51} \\
Large-scale Experiment Support & \ding{55} & \ding{55} & \ding{51} \\
\hline
\end{tabular}
\end{table}

Tab.~\ref{tab:comp} systematically compares representative MCIT toolkits, revealing critical limitations in both quantitative coverage and engineering infrastructure. 
CoIN~\cite{chen2024coin} exhibits a narrow scope, offering only 4 continual learning algorithms and relying on a single benchmark. 
MCITlib~\cite{guo2025mcitlib} further expands this landscape with 8 mainstream algorithms and 3 evaluated datasets; however, both frameworks fundamentally lack a unified backbone and automated support for large-scale experiments. 
Consequently, they often necessitate fragmented configurations, manual intervention, and inconsistent training protocols, which hinder fair cross-method comparison and impede scalable, reproducible research.

To bridge these gaps, we introduce \bibo{}, a plugin-driven reproducible infrastructure specifically designed for scalable MCIT research. It decomposes complex workflows into reusable components for methods, benchmarks, backbones, and evaluation modules, establishing a unified foundation for systematic development. This architecture supports broad algorithmic coverage that encompasses both traditional continual learning baselines and specialized MCIT approaches. By strictly decoupling algorithmic logic from infrastructure maintenance, \bibo{} transforms conventional research pipelines. New methods and benchmarks are integrated as standalone plugins through a lightweight registration mechanism, which isolates implementation details from the underlying MLLM codebase and eliminates structural redundancy. The modular design consolidates training logic into focused wrappers, enabling researchers to inspect and extend algorithms without navigating fragmented repositories. Furthermore, standardized training workflows combined with native support for distributed optimization techniques such as DeepSpeed ensure reproducible experimentation and enable efficient large-scale model training.
Our main contributions are:
\begin{itemize}
    \item A lightweight plugin design that decouples algorithm development from the MLLM backbone, enabling new methods to be integrated with minimal code changes.
    \item A unified benchmarking suite with centralized configuration management, streamlining large-scale experiments and establishing a shared standard for fair method comparison.
\end{itemize}

\begin{figure*}[t]
    \centering
    \includegraphics[width=\linewidth]{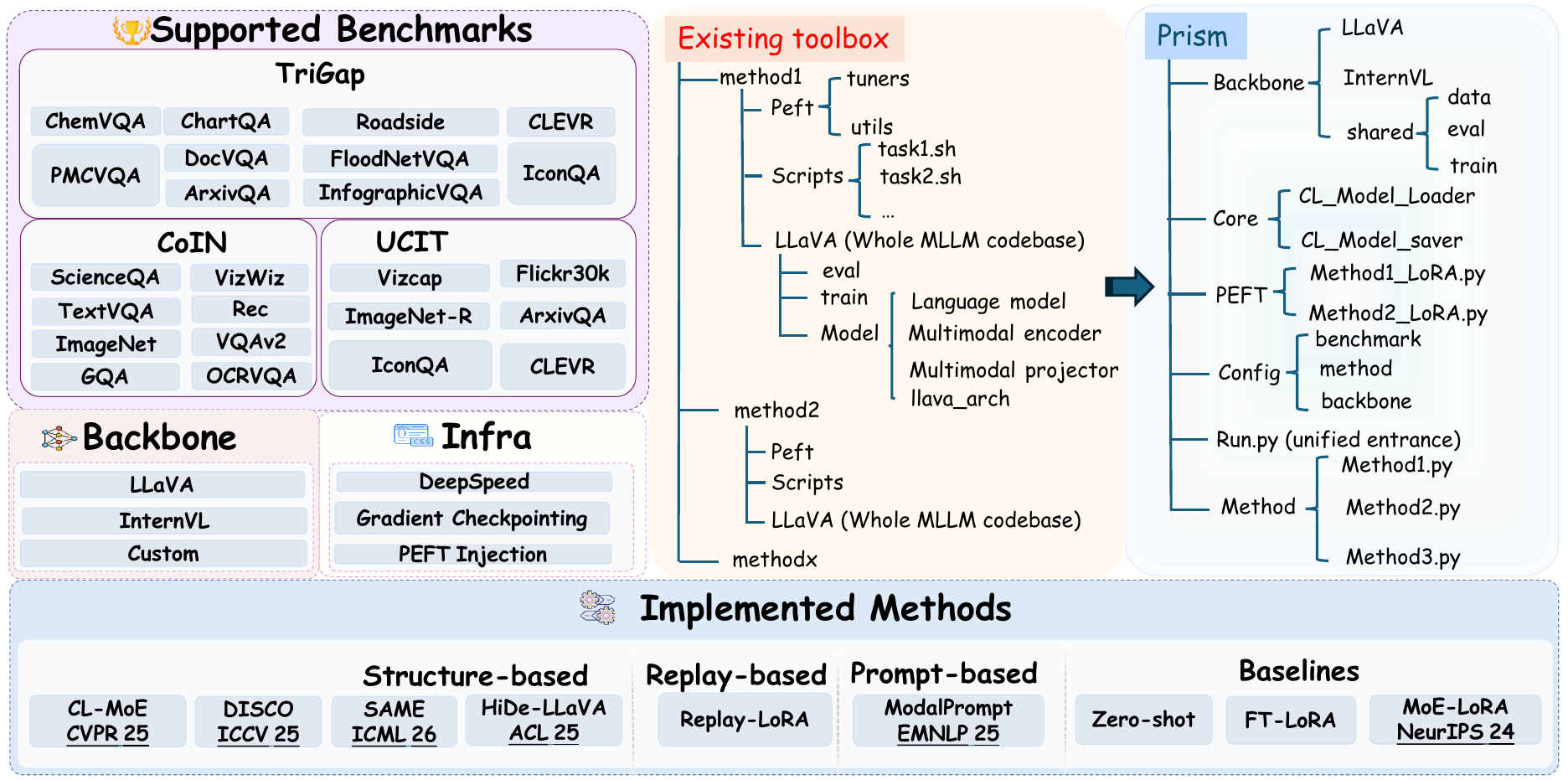}
    \caption{Overview of the \bibo{} toolkit. Its plugin-based design decouples algorithmic development from infrastructure maintenance: new methods, backbones, and benchmarks integrate via lightweight registration, enabling reproducible and extensible MCIT research.}
    \label{fig:teaser}
\end{figure*}

\section{Usage of \bibo}
\textbf{Dependencies.} \bibo\ is built on a modular infrastructure stack for MCIT. The core neural architectures are implemented using PyTorch \cite{paszke2019pytorch} and DeepSpeed \cite{rasley2020deepspeed} for memory-efficient distributed training, HuggingFace Transformers \cite{wolf2020transformers} and PEFT \cite{mangrulkar2022peft} for backbone model management and parameter-efficient fine-tuning, and libraries such as NumPy \cite{harris2020array}, SciPy \cite{virtanen2020scipy}, tqdm \cite{da2019tqdm}, and einops \cite{rogozhnikov2022einops} for numerical operations, monitoring, and tensor manipulation. Notably, our framework is highly extensible and seamlessly supports the integration of multiple custom multimodal backbones such as LLaVA \cite{liu2023llava}, which comprises a CLIP \cite{radford2021learning} vision encoder, a large language model and a visual projector. The project relies solely on widely adopted open-source libraries.

\noindent\textbf{Supported Benchmarks.} We consider 3 benchmarks with diverse domain gaps and task formats, following \cite{guo2025hide,xie2026same}:

\begin{itemize}
    \item \textbf{CoIN} \cite{chen2024coin}: 8 sequential VQA and image understanding tasks: ScienceQA \cite{lu2022learn}, TextVQA \cite{singh2019towards}, ImageNet \cite{deng2009imagenet}, GQA \cite{hudson2019gqa}, VizWiz \cite{gurari2018vizwiz}, Grounding \cite{kazemzadeh2014referitgame,mao2016generation}, VQAv2 \cite{goyal2017making}, and OCRVQA \cite{mishra2019ocr}.
    \item \textbf{UCIT} \cite{guo2025hide}: 6 diverse tasks spanning visual reasoning and captioning: ImageNet-R \cite{hendrycks2021many}, ArxivQA \cite{li2024multimodal}, Vizcap \cite{gurari2018vizwiz}, IconQA \cite{lu2021iconqa}, CLEVR \cite{lindstrom2022clevr}, and Flickr30k \cite{plummer2015flickr30k}.
    \item \textbf{TriGap} \cite{xie2026same}: A long-horizon task sequence consisting of 10 tasks covering document understanding, medical imaging, and domain-specific VQA: PMCVQA \cite{zhang2023pmcvqa}, DocVQA \cite{mathew2020docvqa}, ChartQA \cite{masry-etal-2022-chartqa}, IconQA \cite{lu2021iconqa}, InfographicVQA \cite{mathew2022infographicvqa}, ArxivQA \cite{li2024multimodal}, Roadside \cite{guan2026roadscenevqa}, ChemVQA \cite{sabando2020chemva}, FloodNetVQA \cite{10124393, 9460988}, and CLEVR \cite{lindstrom2022clevr}.
\end{itemize}
\vspace{-3mm}
\noindent\textbf{Task Organization.} Following the protocols in continual instruction tuning \cite{chen2024coin,guo2025hide}, \bibo\ organizes tasks sequentially. Each benchmark defines a fixed task order, where the model incrementally learns each task.

\noindent\textbf{Implemented Methods.} \bibo\ implements a total of 9 representative continual learning methods and baselines for multimodal LLMs. These are systematically categorized into: 
(1) Baselines, which establish performance boundaries for evaluation, including \texttt{Zero-shot} (Zero-shot LLaVA without any fine-tuning), \texttt{FT-LoRA} (sequential full LoRA fine-tuning representing catastrophic forgetting), and \texttt{MoE-LoRA} \cite{chen2024coin};
(2) Structure-based methods, which mitigate forgetting via explicit parameter isolation or routing, covering \texttt{HiDe-LLaVA} \cite{guo2025hide}, \texttt{DISCO} \cite{guo2025federated}, \texttt{CL-MoE} \cite{huai2025cl}, and \texttt{SAME} \cite{xie2026same}; 
(3) Replay-based methods, \textit{i.e.}, \texttt{Replay-LoRA} (LoRA with task-partitioned experience replay); and 
(4) Prompt-based methods, \textit{i.e.}, \texttt{ModalPrompt} \cite{zeng2025modalprompt}. 
All methods share a unified PEFT injection interface; new methods are seamlessly added via \texttt{method/<name>/integration.py} and registered with \texttt{@CLMethodFactory.register()}.

\begin{table*}[t]
\centering
\small
\setlength{\extrarowheight}{1pt}
\setlength{\tabcolsep}{17pt}
\renewcommand{\arraystretch}{1.0}
\caption{Average performance of different methods on the UCIT benchmark. The best and second-best results are highlighted in \textbf{bold} and \underline{underline}, respectively.}
\label{tab:ucit}
\resizebox{\linewidth}{!}{
\begin{tabular}{l|cccccc|c}
\hline
\rowcolor{gray!20}
\textbf{Methods} & ImageNet-R & ArxivQA & Vizcap & IconQA & CLEVER & Flickr30k & Average \\
\hline

Zero-shot & 18.88 & 52.62 & 38.75 & 21.25 & 21.12 & 41.44 & 32.34 \\
\rowcolor{gray!10}
FT-LoRA & 29.33 & 55.30 & 45.51 & 26.13 & 13.07 & \underline{58.07} & 37.90 \\

Replay-LoRA & 76.93 & 87.07 & 54.31 & 56.43 & 36.40 & 55.94 & 61.18 \\
\rowcolor{gray!10}
MoE-LoRA~\citep{chen2024coin} & 58.43 & 77.57 & 44.83 & 68.90 & 56.73 & \textbf{58.27} & 60.79 \\

HiDe-LLaVA~\citep{guo2025hide} & 87.62 & 91.12 & 42.68 & 57.62 & 31.00 & 50.41 & 60.08 \\
\rowcolor{gray!10}
ModalPrompt~\citep{zeng2025modalprompt} & 80.50 & 90.62 & \textbf{60.13} & 63.50 & 55.75 & 57.09 & 67.93 \\
CL-MoE~\citep{huai2025cl} & 64.12 & 78.38 & 44.83 & 62.00 & 50.75 & 58.06 & 59.69 \\
\rowcolor{gray!10}
DISCO~\citep{guo2025federated} & \underline{88.88} & \textbf{94.25} & 47.52 & \underline{69.50} & \underline{60.75} & 56.32 & \underline{69.54} \\
SAME~\citep{xie2026same} & \textbf{89.91} & \underline{91.40} & \underline{55.33} & \textbf{77.51} & \textbf{68.85} & 55.43 & \textbf{73.07} \\
\hline
\end{tabular}
}
\end{table*}

\begin{table*}[t]
\centering
\renewcommand{\arraystretch}{1.0}
\setlength{\extrarowheight}{1pt}
\caption{Average performance of different methods on TriGap benchmark. The best and second-best results are highlighted in \textbf{bold} and \underline{underline}, respectively.}
\label{tab:main}
\resizebox{\linewidth}{!}{
\begin{tabular}{l|cccccccccc|c}
\hline
\rowcolor{gray!20}
\textbf{Methods} & PMCVQA & DocVQA & ChartQA & IconQA & InfographicVQA & ArxivQA & Roadside & ChemVQA & FloodNetVQA & CLEVR & Average \\
\hline

Zero-shot & 35.40 & 12.68 & 9.36 & 19.27 & 5.06 & 53.77 & 7.40 & 5.30 & 47.41 & 20.37 & 21.60 \\
\rowcolor{gray!10}
FT-LoRA & 34.20 & 23.32 & 9.84 & 37.07 & 23.53 & 83.83 & 7.00 & 12.70 & 80.31 & 60.27 & 37.21 \\
Replay-LoRA & 33.70 & 33.95 & 14.00 & 46.67 & 28.97 & 75.57 & 9.40 & 15.90 & 73.81 & 58.80 & 39.08 \\
\rowcolor{gray!10}
MoE-LoRA~\citep{chen2024coin} & 39.03 & 37.49 & 12.44 & 43.43 & 35.17 & 90.90 & 7.93 & 20.70 & \textbf{90.41} & \textbf{67.00} & 44.45 \\
HiDe-LLaVA~\citep{guo2025hide} & 37.00 & 33.20 & 10.52 & 41.97 & 24.09 & 79.20 & 7.73 & 11.17 & 57.39 & 23.00 & 32.53 \\
\rowcolor{gray!10}
ModalPrompt~\citep{zeng2025modalprompt} & 38.23 & 38.23 & 11.92 & 44.73 & 37.37 & 84.47 & 10.13 & 12.43 & 71.52 & 52.50 & 40.15 \\
CL-MoE~\citep{huai2025cl} & 40.53 & 36.79 & 13.72 & 52.70 & 32.27 & \textbf{93.00} & 7.77 & 18.33 & 80.09 & \underline{65.90} & 44.11 \\
\rowcolor{gray!10}
DISCO~\citep{guo2025federated} & \textbf{42.03} & \underline{43.50} & \textbf{18.01} & \underline{63.13} & \underline{38.23} & \underline{91.27} & \textbf{11.02} & \textbf{22.13} & 80.25 & 55.87 & \textbf{46.54} \\
SAME~\citep{xie2026same} & \underline{41.60} & \textbf{43.87} & \underline{17.56} & \textbf{64.03} & \textbf{39.57} & 90.46 & \underline{10.83} & \underline{21.77} & \underline{81.09} & 54.50 & \underline{46.53} \\

\hline
\end{tabular}}
\end{table*}

\noindent\textbf{Evaluation Metrics.} Following standard continual learning evaluation protocols \cite{zhou2024class,guo2025hide}, we denote $A_t$ as the model's accuracy after the $t$-th incremental stage. \bibo\ employs the following primary metrics:

\begin{itemize}
\vspace{-2.8mm}
    \item \textbf{Last Accuracy} $A_B$: performance after the final task.
    \vspace{-2.8mm}
    \item \textbf{Average Accuracy} $\bar{A} = \frac{1}{T}\sum_{t=1}^{T} A_t$: mean accuracy across all incremental stages.
    \vspace{-2.8mm}
    \item \textbf{Forgetting Measure}: $F_T$ is utilized to  measure the average performance drop of each task from its best-achieved accuracy to the final stage, \textit{i.e.,} $F_T = \frac{1}{T-1}\sum_{t=1}^{T-1}\max_{t\le l\le T-1}(A_{l,t} - A_{T,t})$
\end{itemize}
For VQA tasks (\textit{e.g.}, VQAv2, TextVQA, GQA, VizWiz, ScienceQA), accuracy is computed via string-matching with normalization following the standard VQA evaluation protocol \cite{antol2015vqa}. For captioning tasks (\textit{e.g.}, Flickr30k, Vizcap), standard COCO metrics, including CIDEr \cite{vedantam2015cider}, BLEU \cite{papineni2002bleu}, METEOR \cite{banerjee2005meteor}, ROUGE-L \cite{lin2004rouge}, SPICE \cite{anderson2016spice}) are employed. For classification-style tasks (\textit{e.g.}, ImageNet-R, ArxivQA, IconQA, CLEVR), exact-match accuracy is used.

\noindent\textbf{Basic Usage.} \bibo\ centralizes all experimental parameters (benchmarks, methods, training protocols) in human-readable Python configuration files, eliminating the need to modify underlying code. Users can simply adjust parameters within the configuration files and run standardized commands as:
\begin{verbatim}
python run.py {train|infer} <task_ids> \
--benchmark <benchmark> --method <method>
\end{verbatim}
where \texttt{<benchmark>} is one of the supported benchmarks; \texttt{<method>} corresponds to one of the implemented methods; and \texttt{<task\_ids>} specifies the sequential task indices to run.

\noindent\textbf{Configuration.} All experimental settings and parameters are centralized in a modular configuration system. For a detailed breakdown of the configuration files and directory structure (covering methods, benchmarks, backbones, and DeepSpeed settings), please refer to Appendix~\ref{app:configuration}.

\section{Experiment}
We evaluate all methods on UCIT and TriGap using the LLaVA-v1.5-7B backbone, trained on 4 NVIDIA RTX 5090 GPUs. Comprehensive results are summarized in Tab.~\ref{tab:ucit} and~\ref{tab:main}, with detailed implementation settings provided in Appendix~\ref{details}.

\noindent Overall, the baselines establish clear performance boundaries: Zero-shot serves as a reference for initial capability, while FT-LoRA and MoE-LoRA illustrate typical catastrophic forgetting patterns. Among MCIT strategies, structure-based methods demonstrate the strongest performance through parameter isolation and expert routing. The replay-based approach ensures memory retention via historical data rehearsal. Furthermore, while prompt-based methods minimize trainable parameters, they require significantly more training epochs to converge, resulting in prolonged training time. Beyond these category-specific trends, we observe substantial performance fluctuations across benchmarks. Notably, on a highly challenging benchmark such as TriGap, the amount of parameters allocated per task significantly impacts final accuracy.

\section{Conclusion}
In this paper, we introduce \bibo, a plugin-extensible toolbox that lowers the engineering barrier in multimodal continual instruction tuning. By decoupling algorithm development from infrastructure via lightweight registration, \bibo\ enables researchers to implement and reproduce methods by modifying a minimal amount of code. \bibo\ establishes a shared infrastructure for reproducible, extensible, and scalable MCIT research.

\noindent\textbf{Limitations.}
\bibo\ does not currently cover all MCIT methods and MLLM backbones. However, its plugin-centric architecture inherently streamlines the integration of new algorithms. Extending this coverage to a broader range of methods and MLLM families remains future work.

\bibliography{custom}

\appendix
\clearpage

\section{Brief Introduction of Reproduced Methods}
\label{sec: Brief Introduction of Reproduced Methods}

\noindent\textbf{Zero-shot.} A baseline that evaluates the frozen pre-trained LLaVA model on all tasks without any fine-tuning, measuring the inherent zero-shot generalization of the multimodal backbone.

\noindent\textbf{FT-LoRA.} A sequential LoRA fine-tuning baseline that injects trainable low-rank adapters into the LLM backbone. Each task is trained sequentially, with only the LoRA parameters updated while the base model remains frozen.

\noindent\textbf{Replay-LoRA.} A replay-assisted LoRA method that maintains a task-partitioned memory buffer of training examples from previous tasks. During each training step, stored examples are sampled and replayed alongside the current task data to reinforce prior knowledge.

\noindent\textbf{MoE-LoRA} \cite{chen2024coin}. A mixture-of-experts LoRA variant that introduces multiple expert LoRA groups per layer with a learned soft router. The router produces a weighted combination of expert outputs, enabling the model to dynamically allocate capacity across tasks.

\noindent\textbf{HiDe-LLaVA} \cite{guo2025hide}. A HiDe-style mixture-of-experts LoRA approach that maintains per-layer task-specific expert LoRA groups. During training, only the expert corresponding to the current task is activated; during inference, task identity is inferred via CLIP-based image and text anchor matching to route to the appropriate expert.

\noindent\textbf{CL-MoE} \cite{huai2025cl}. A continual learning mixture-of-experts method using input-dependent per-layer per-token routing, eliminating the need for explicit task-ID gating. Combined with memory replay, it provides a strong task-agnostic baseline for continual instruction tuning.

\noindent\textbf{DISCO} \cite{guo2025federated}. A diagonal mask routing MoE-LoRA approach that learns per-task CLIP-based image and text prototypes. During inference, cosine similarity between the input features and stored prototypes produces diagonal mask weights for expert aggregation, enabling task-identity-aware routing without explicit task IDs.

\noindent\textbf{ModalPrompt} \cite{zeng2025modalprompt}. A prompt-based method that learns per-task soft prompts prepended to the input embedding sequence. At inference, dual-modal guidance is used to select the top-$K$ most relevant prompts, with a tunable balance parameter $\lambda$ controlling the image-text mixing weight.

\noindent\textbf{SAME} \cite{xie2026same}. A spectral anchor-based method that performs online SVD of the covariance matrix of LoRA parameters within a sliding window. The principal singular vectors are retained as task-anchoring directions, and a curvature-aware importance score guides parameter consolidation across tasks.

\begin{table*}[t]
\centering
\caption{Details of datasets used in UCIT benchmark.}
\label{tab:ucit_details}
\setlength{\tabcolsep}{8pt}
\renewcommand{\arraystretch}{1.2}
\begin{tabular}{l c c l}
\textbf{Dataset} & \textbf{Train} & \textbf{Test} & \textbf{Domain Description} \\
ImageNet-R & 24000 & 3000 & Object recognition with artistic renditions \\
ArxivQA    & 40000 & 3000 & Academic paper figure understanding \\
VizWiz     & 40000 & 3000 & Visual assistance for visually impaired \\
IconQA     & 30000 & 3000 & Icon comprehension \\
CLEVR-Math & 40000 & 3000 & Mathematical reasoning on synthetic scenes \\
Flickr30k  & 40000 & 3000 & Image captioning for real-world photos \\
\end{tabular}
\end{table*}

\begin{table*}[t]
\centering
\caption{Details of datasets used in TriGap benchmark.}
\label{tab:trigap_details}
\setlength{\tabcolsep}{8pt}
\renewcommand{\arraystretch}{1.2}
\begin{tabular}{l c c l}
\textbf{Dataset} & \textbf{Train} & \textbf{Test} & \textbf{Domain Description} \\
PMCVQA         & 40000 & 3000 & Medical image analysis and diagnosis \\
DocVQA         & 30000 & 3000 & Document understanding and text extraction \\
ChartQA        & 25000 & 3000 & Chart and graph reasoning \\
IconQA         & 10000 & 3000 & Icon comprehension \\
InfographicVQA & 20000 & 3000 & Infographic information extraction \\
ArxivQA        & 10000 & 3000 & Academic paper figure analysis \\
Roadside       & 40000 & 3000 & Autonomous driving scene understanding \\
ChemVQA        & 40000 & 3000 & Molecular structure analysis \\
FloodNetVQA    & 10000 & 3000 & Disaster scene assessment \\
CLEVR          & 10000 & 3000 & Mathematical reasoning on synthetic scenes \\
\end{tabular}
\end{table*}

\section{Brief Introduction of Selected Benchmarks}
Tables~\ref{tab:ucit_details} and~\ref{tab:trigap_details} summarize the dataset compositions of the UCIT~\cite{guo2025hide} and TriGap~\cite{xie2026same} benchmarks, respectively. Both benchmarks strictly enforce an \textit{unseen-data} protocol: all samples are rigorously filtered to ensure zero overlap with the pre-training or supervised fine-tuning (SFT) corpora of the underlying MLLMs, thereby eliminating potential information leakage and guaranteeing fair evaluation of continual learning capabilities. UCIT comprises six tasks with training sets ranging from 24k to 40k samples and a uniform test split of 3k per task, offering a lightweight and standardized protocol for efficient method validation. In contrast, TriGap expands the scope to ten highly heterogeneous domains, with training sizes varying from 10k to 40k to reflect real-world data availability across specialized fields (\textit{e.g.}, medical imaging, autonomous driving, chemical analysis). By maximizing both the task sequence length and inter-domain distribution shifts, TriGap serves as a comprehensive, high-difficulty benchmark designed for stress-testing long-term knowledge retention. Together, these two benchmarks form a complementary evaluation suite: UCIT provides a controlled baseline, while TriGap offers a rigorous, large-scale setting for assessing model robustness and anti-forgetting capabilities under extreme distribution shifts.

\section{Implementation Details}
\label{details}
All methods are built upon the LLaVA-1.5 architecture, which consists of a Vicuna-7B LLM backbone and a CLIP-ViT-L/14 vision encoder.
Unless otherwise noted, all methods share the following training configuration: AdamW optimizer with learning rate $2\times10^{-4}$, cosine schedule with 3\% warmup, weight decay 0.0, bf16 mixed precision, model max length 2048, gradient checkpointing enabled, and 1 training epoch.
All adapter modules are injected exclusively into the LLM backbone, with LoRA target modules and rank configurations for select methods adopted directly from their official implementations.

\subsection{Zero-shot}
Zero-shot serves as a parameter-free baseline that bypasses continual instruction tuning entirely.

\noindent\textbf{Insertion.} No PEFT modules or task-specific adapters are injected. The model operates directly on the frozen pretrained MLLM weights without any parameter updates or checkpoint loading throughout the continual learning sequence.

\noindent\textbf{Hyperparameters.} As an inference-only baseline, Zero-shot is excluded from the training pipeline. Evaluation adopts the standard decoding configuration (\textit{e.g.}, conversation template and temperature) shared across all methods.

\subsection{FT-LoRA}
FT-LoRA is a sequential fine-tuning baseline that applies standard LoRA adapters without continual learning mechanisms.

\noindent\textbf{Insertion.} Standard LoRA adapters are injected into the attention and FFN linear layers ($q_{\mathrm{proj}}$, $k_{\mathrm{proj}}$, $v_{\mathrm{proj}}$, $o_{\mathrm{proj}}$, $\mathrm{gate}_{\mathrm{proj}}$, $\mathrm{up}_{\mathrm{proj}}$, $\mathrm{down}_{\mathrm{proj}}$) of the LLM trunk. The vision tower and multimodal projector remain frozen.

\noindent\textbf{Hyperparameters.} We set the LoRA rank and scaling factor as $r=96, \alpha=192$ for UCIT; and $r=80, \alpha=160$ for TriGap. LoRA dropout is fixed at 0.05. Training runs for 1 epoch with a learning rate of $2\times10^{-4}$ (cosine schedule, warmup ratio 0.03) and a projector learning rate of $2\times10^{-5}$. Per-device batch size is 12 for all tasks on CoIN and UCIT.

\subsection{Replay-LoRA}
Replay-LoRA extends FT-LoRA by incorporating a task-partitioned experience replay buffer to mitigate catastrophic forgetting.

\noindent\textbf{Insertion.} The adapter insertion follows FT-LoRA (LoRA on attention and FFN layers of the LLM trunk). Additionally, a task-partitioned replay buffer stores samples from previous tasks. During training on task $t$, historical samples are merged into the current dataloader via a replay-sidecar JSON configuration.

\noindent\textbf{Hyperparameters.} LoRA configurations match FT-LoRA ($r=96, \alpha=192$ for UCIT; $r=80, \alpha=160$ for TriGap; dropout 0.05; 1 epoch; LR $2\times10^{-4}$; projector LR $2\times10^{-5}$). Replay-specific settings include a total buffer capacity of 180 samples (evenly distributed across the first $N-1$ tasks) and a per-example sampling probability of 0.7. Per-device batch sizes are 12.

\subsection{HiDe-LLaVA}
HiDe-LLaVA introduces a hierarchical decoupling mechanism with task-specific LoRA experts and dual-modal prototype routing.

\noindent\textbf{Insertion.} The attention and FFN layers are replaced with \texttt{HiDeMOELoraLinear} modules. Each layer hosts $N$ task-specific LoRA experts (where $N$ is the total number of tasks) alongside a lightweight per-layer router. Adapters are applied exclusively to the LLM, while frozen CLIP-derived image and text anchors are stored per task for inference-time routing.

\noindent\textbf{Hyperparameters.} LoRA settings are $r=96, \alpha=192$ (UCIT), and $r=80, \alpha=160$ (TriGap), with dropout 0.05. Training uses 1 epoch, LR $2\times10^{-4}$, and projector LR $2\times10^{-5}$. The CLIP feature dimension is 768 (CLIP-ViT-L/14). Per-device batch sizes are 12.

\noindent\textbf{Routing.} During training, the active expert is selected via the current task ID. At inference, dual-modal prototype matching assigns the task: image and text features are compared to per-task anchors using cosine similarity, combined as $0.5 \cdot \mathrm{sim}_{\mathrm{image}} + 0.5 \cdot \mathrm{sim}_{\mathrm{text}}$, with the argmax index yielding the \texttt{predicted\_task\_id}. On the final transformer block, only the predicted expert is activated; on preceding blocks, LoRA deltas from all experts are fused via summation. Text-only inputs default to text-anchor matching.

\subsection{CL-MoE}

\textbf{Insertion.} CL-MoE replaces all FFN linear layers ($\mathrm{gate}_{\mathrm{proj}}$, $\mathrm{up}_{\mathrm{proj}}$, $\mathrm{down}_{\mathrm{proj}}$) with \texttt{CLMoELinear} modules.
Each layer contains $N$ independent LoRA expert branches, where $N$ is the total number of tasks.
The total LoRA rank is evenly split across experts (per-expert rank $= r / N$).

\noindent\textbf{Hyperparameters.} We use $r=96, \alpha=192$ for UCIT; and $r=80, \alpha=160$ for TriGap.
LoRA dropout is set to 0.05.
The task embedding dimension is 64.
Training uses a per-device batch size of 4 across all benchmarks.

\subsection{DiSCO}

\textbf{Insertion.} DiSCO replaces the same set of FFN linear layers as CL-MoE with \texttt{DiscoMOELoraLinear} modules.
The LoRA rank is adjusted to be divisible by the number of tasks, and $\alpha$ is set to $2 \times \mathrm{adjusted\_}r$.

\noindent\textbf{Hyperparameters.} We use $r=96, \alpha=192$ for UCIT; and $r=80, \alpha=160$ for TriGap.
LoRA dropout is 0.05.
The routing temperature $\tau$ is set to 0.05, and the CLIP feature dimension is 768 (matching CLIP-ViT-L/14).
Training uses a per-device batch size of 4 for all benchmarks.

\subsection{ModalPrompt}

ModalPrompt does \emph{not} use LoRA adapters. Instead, it introduces per-task learnable soft prompt tokens and prompt transformation MLPs.

\noindent\textbf{Insertion.} Soft prompts are prepended to the input sequence at the embedding level.
Each task is assigned a learnable prompt of $\mathrm{prefix\_len}=10$ continuous tokens and a dedicated prompt transform MLP that maps the prompt into a feature space aligned with CLIP representations.
The transformation MLP is trained via a cosine similarity loss against the corresponding CLIP features.

\noindent\textbf{Hyperparameters.} The number of top-$K$ prompts selected per inference step is $\mathrm{transfer\_num}=1$.
The dual-modal guidance coefficient $\lambda$ is set to 0.5, balancing image and text prototype similarities as $\lambda \cdot \mathrm{sim}_{\mathrm{image}} + (1-\lambda) \cdot \mathrm{sim}_{\mathrm{text}}$.
The prototype momentum for EMA updates is 0.9.
ModalPrompt is trained for 4 epochs.
Per-device batch sizes are 4 for all benchmarks.

\subsection{MoE-LoRA}

\textbf{Insertion.} MoE-LoRA replaces FFN linear layers ($\mathrm{gate}_{\mathrm{proj}}$, $\mathrm{up}_{\mathrm{proj}}$, $\mathrm{down}_{\mathrm{proj}}$) with \texttt{MoELoRALinear} modules.
The total LoRA rank $r$ must be divisible by the number of experts $N$, with each expert receiving rank $r / N$.

\noindent\textbf{Hyperparameters.} We use rank $r=96, \alpha=192$ for UCIT; and $r=80, \alpha=160$ for TriGap.
LoRA dropout is 0.05.
Per-device batch sizes are 4 for all benchmarks.

\subsection{SAME}

\textbf{Insertion.} SAME replaces only FFN linear layers ($\mathrm{gate}_{\mathrm{proj}}$, $\mathrm{up}_{\mathrm{proj}}$, $\mathrm{down}_{\mathrm{proj}}$) with \texttt{SAMELinear} modules.
Each layer maintains per-task LoRA expert weights along with task-wise covariance matrices for spectral analysis and parameter sharing.

\noindent\textbf{Hyperparameters.} We use $r=96, \alpha=192$ for UCIT; and $r=80, \alpha=160$ for TriGap.
LoRA dropout is 0.05.
SAME-specific hyperparameters include: the curvature/saliency threshold $\tau_{\mathrm{score}}=0.1$, curvature EMA momentum $\mu=0.9$, curvature estimation window size 3, maximum number of principal components 64, and cumulative energy ratio 0.9 for SVD-based truncation.
Per-device batch sizes are 4 for all benchmarks.

\section{Configuration Details}
\label{app:configuration}

All experimental settings and parameters in our framework are centralized and can be configured in the following files and directories:
\begin{itemize}
    \item \texttt{config/run\_config.py}: Defines global CLI arguments for training/inference, covering benchmark, method, and GPU allocation.
    \item \texttt{config/methods/}: Method-specific hyperparameter configurations.
    \item \texttt{config/benchmarks/}: Benchmark-specific configurations, including task definitions, dataset paths, and evaluation hooks.
    \item \texttt{config/backbone/}: Backbone identifier and default conversation template.
    \item \texttt{config/paths/}: Filesystem paths for model weights, datasets and checkpoints.
    \item \texttt{config/deepspeed/}: DeepSpeed ZeRO configuration files (stage 2, 3, and 3 offload).
\end{itemize}

\label{sec:methods}

\end{document}